\documentclass[times, twoside]{arxivStyle}
\usepackage{blindtext}

\usepackage[disable]{todonotes}
\usepackage{verbatim}
\usepackage{multicol}
\usepackage{pdflscape}
\usepackage{makecell}
\usepackage{adjustbox}
\usepackage{float}

\usepackage{textcomp,mathcomp,amsmath}

\leadauthor{Albrecht, Godinho} 

\begin{document}




\title{\huge ARI3D: A Software for Interactive Quantification of Regions in X-Ray CT 3D Images}

\shorttitle{ARI3D}

\author[1,2,3,$\ast$,\Letter]{Jan Phillipp Albrecht}
\author[4,$\ast$,$\dagger$,\Letter]{Jose R.A. Godinho}
\author[1,2]{Christina Hübers}
\author[1,2,$\dagger$,\Letter]{Deborah Schmidt}

\affil[1]{Max Delbr\"uck Center for Molecular Medicine in the Helmholtz Association, Berlin, Germany}
\affil[2]{
Helmholtz Imaging, Max Delbr\"uck Center for Molecular Medicine in the Helmholtz Association, Berlin, Germany}
\affil[3]{Humboldt-Universität zu Berlin, Faculty of Mathematics and Natural Sciences, Berlin, Germany}
\affil[4]{Helmholtz-Zentrum Dresden-Rossendorf, Helmholtz-Institut Freiberg für Ressourcentechnologie (HIF), Berlin, Germany}

\affil[$\ast$]{These authors contributed equally to this work.} 
\affil[$\dagger$]{These authors jointly supervised this work.}
\affil[]{Correspondence: jan-philipp.albrecht@mdc-berlin.de, jragodinho@uma.es, deborah.schmidt@mdc-berlin.de}

\maketitle






\begin{abstract}
X-ray computed tomography (CT) is the main 3D technique for imaging the internal microstructures of materials. Quantitative analysis of the microstructures is usually achieved by applying a sequence of steps that are implemented to the entire 3D image. This is challenged by various imaging artifacts inherent from the technique, e.g., beam hardening and partial volume. Consequently, the analysis requires users to make a number of decisions to segment and classify the microstructures based on the voxel gray-values. In this context, a software tool, here called ARI3D, is proposed to interactively analyze regions in three-dimensional X-ray CT images, assisting users through the various steps of a protocol designed to classify and quantify objects within regions of a three-dimensional image. ARI3D aims to 1) Improve phase identification; 2) Account for partial volume effect; 3) Increase the detection limit and accuracy of object quantification; and 4) Harmonize quantitative 3D analysis that can be implemented in different fields of science.

\end{abstract}


\section*{Motivation and significance}
X-ray computed tomography (CT) utilizes the penetrative nature of high energy X-rays to create 3D images of the internal components of samples (also referred to as materials or phases depending on the scientific field) \cite{cnudde_high-resolution_2013, withers_x-ray_2021}. Despite enabling unique insights, interpreting data from an X-ray beam that has crossed different sample components is challenging. For example, contrast and spatial resolution may be compromised due to imaging artifacts arising from the polychromatic energy spectrum \cite{bultreys_fast_2016, wensink_spontaneous_2023} and the geometry of the beam \cite{gajjar_new_2018, godinho_spectral_2021}. The resulting image artifacts unpredictably affect the gray-values of the voxels. Since the artifacts are inevitable, independently of the scientific field, it is difficult to develop standardized solutions to correct for these artifacts that work across applications.
The image artifacts pose challenges during segmentation and classification of objects based on voxel gray-values. Segmentation of a grayscale image consists of defining the voxelized space representing an object by binarizing the image \cite{buyse_combining_2023}, i.e., each voxel is either a part or excluded from the object(s). Classification entails identifying the material(s) within the segmented region \cite{wang_current_2020} also utilizing gray-values. The problem is that the exact boundary of irregular objects cannot fit perfectly on a voxelized grid that constitutes the image resulting in many voxels containing only a partial volume of the object. In practice, due to cone beam artifacts, the partial volume effect is blurred across 3 to 7 voxel layers causing a gradient of gray-values at the interphase \cite{zwanenburg_effective_2023}. If the composition in both sides of an interphase are known, the partial volume can be accounted for without segmentation of the two objects \cite{zwanenburg_performance_2022, godinho_volume_2019}. Nevertheless, in complex samples with multiple objects and interphases it is difficult to apply such methods.  
To simplify their application, the 3D image can be split into less complex regions, each with fewer objects and interphases. This has been demonstrated by applying the MSPaCMAn workflow \cite{godinho_mounted_2021, godinho_3d_2023, gupta_standardized_2024} that was developed to analyse particulate materials, whereby each region is one particle with a unique set of properties. So far, the workflow included several python scripts and required other software to view images and plot data, which reduced its usability by the wider community. Here, a software is proposed to 1) guide end-users without coding knowledge through the workflow; 2) facilitate future software updates of individual steps of the workflow without altering the core components of the code due the software modular design.

\begin{figure*}[tp!]
    \centering
    \includegraphics[width=\textwidth]{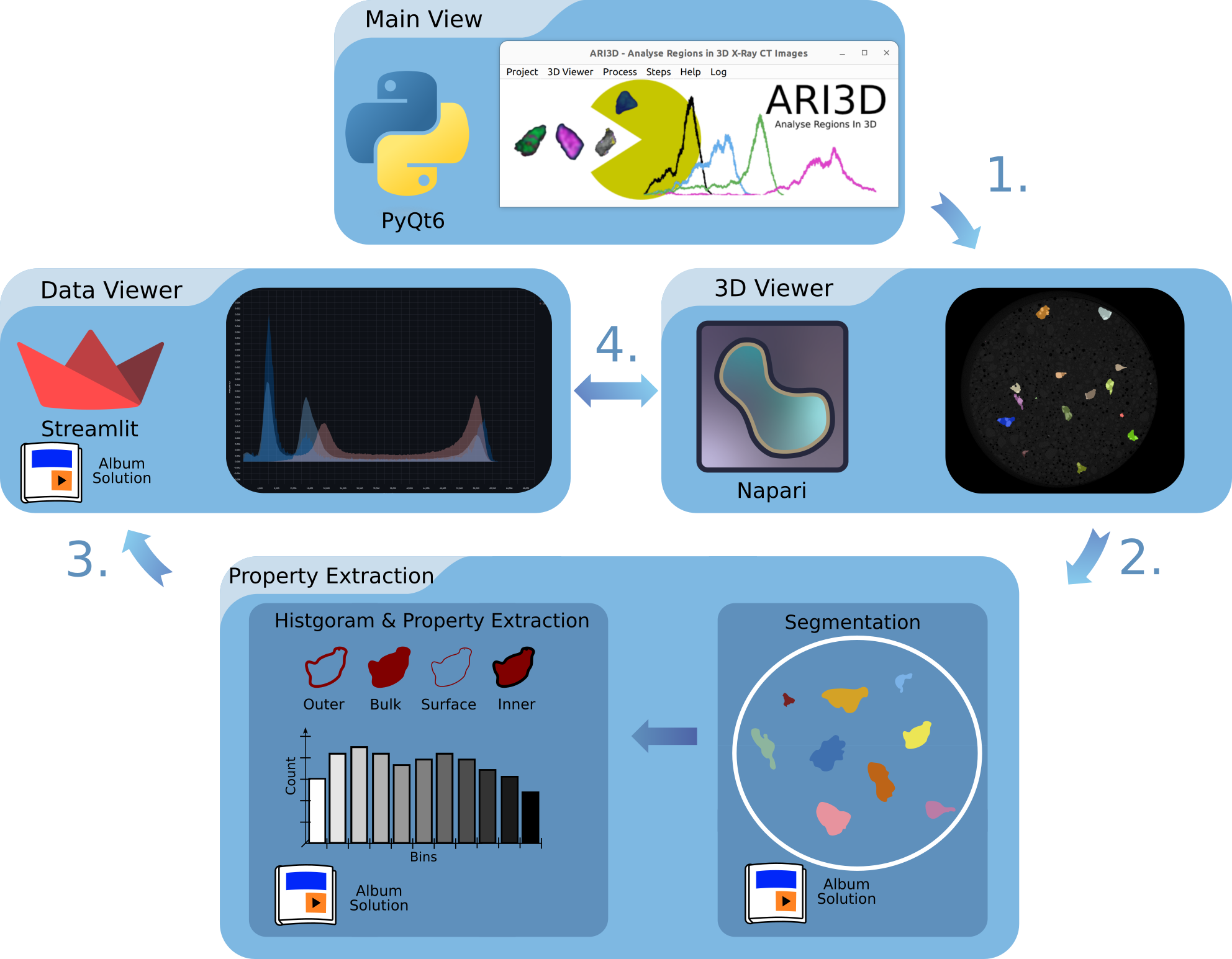}
    \caption{Schematic view of the software components Main Window, 3D Viewer, property extraction and Data Viewer. All components can be initiated via the Main Window. Arrows with numbers indicate the order in which the software components are connected in the workflow. Double ended arrows point to components interacting with each other. For a more detailed description of the entire workflow we refer to Annex A and Annex B. Property extraction and the data viewer are released with album in the form of solutions (symbolized by the white book icon).
}
    \label{fig:schamtic_view}
\end{figure*}

\section*{Software description}
The software has been designed to integrate the primary steps of the MSPaCMAn workflow \cite{gupta_standardized_2024}. First, the 3D image is segmented into subregions, and the unique properties of each region are extracted. Second, these properties are analyzed and interpreted to identify and quantify the materials that compose each region (a detailed description is provided in Annex \ref{annex:workflow}). Throughout the workflow, the user must evaluate various parameters specific to a sample and specify various inputs accordingly (see Annex \ref{annex:user_parameters}). Consequently, effective utilization of the workflow necessitates a graphical user interface that allows users to interact with the data corresponding to individual and multiple regions at various stages of the workflow.

\subsection*{Software architecture}
The software comprises four modular interconnected components: i) the main window, ii) 3D viewer, iii) Property Extraction, and iv) data viewer, as illustrated in Figure 1. 

The main window was developed using pyQt to facilitate access to the primary functionalities necessary to implement the workflow and access project-level features. These functionalities can operate independently or as a pipeline by importing input parameters from a settings file. There are six main tabs available: the Project tab, which enables project-level control; the 3D Viewer tab, which provides access to image operations; the Process tab, which allows for region extraction operations; the Steps tab, which enables control over the modular components; the Help tab, which contains links to step-by-step instructions and tutorials; and the Log tab, which allows toggling verbosity to assist future developers.  

The 3D viewer utilizes a customized version of the open-source library Napari \cite{sofroniew_napari_2025}. This tool facilitates interactive visualization of 3D images, offering various options for image manipulation, such as rotation, transparency adjustment, and contrast enhancement. Customized widgets provide the basic segmentation functionalities that are essential to the workflow, as detailed in Annex \ref{annex:workflow}.

Property extraction is the process of extracting information from the image for the downstream analysis. This includes the geometric and grayscale properties of individual regions. An optional segmentation approach using a deep learning segmentation tool is included specifically for particulate materials (ParticleSeg 3D) \cite{gotkowski_particleseg3d_2024}. Note that the segmentation and labeling of the regions can be done in any other software, although the interactive connection with the DataViewer (explained in 2.2) requires loading a labeled image. Both the deep learning segmentation and property extraction steps are realized as a modular component.  

The data viewer is designed as a modular component and built on the Streamlit \footnote{https://streamlit.io/} app framework for interactive data analysis. It enables the visualization of histograms and properties of individual regions that were extracted from the 3D image. Various plots are generated using the Vega-Altair declarative library \cite{vanderplas_altair_2018} and are interactively linked to input widgets, such as buttons and sliders. This functionality allows for near real-time testing of how changes in input parameters affect the quantification of large amounts of regions. The distribution of tabs facilitates the progression through the workflow steps. Importantly, the 3D image of a region and its corresponding histogram can be visualized live by a simple click of the mouse. Therefore, the various input parameters can be validated by the user by confronting the result of the quantification in a region with the image of that region, before the workflow is applied to all regions.
The modularity of the components for segmentation, property extraction, and the data viewer is realized via album \cite{albrecht_album_2021}. Album offers the capability to execute code within its respective virtual environment, referred to as a solution. Our software utilizes this feature to enable modularity for the workflow components. Moreover, by utilizing Album, solutions can be individually updated by future developers and reinstalled by users through the main menu of our software. For more details about modularity, project folders, files, and code structure, we refer to Annex \ref{annex:code_structure}.

\subsection*{Software functionalities}
The software implements the main functionalities of the workflow necessary to go from a grayscale image (input) to a csv containing a list of properties of all regions in a sample (output). See description in Annex \ref{annex:workflow} and diagram in Annex \ref{annex:project_structure}. 

\subsubsection*{Project Management}
We provide project management functionality within the primary interface of our software. This feature allows for the loading and saving of projects, which enables users to resume work on a previously saved project. FAIR \cite{wilkinson_fair_2016, barker_introducing_2022} practices are promoted by the storage of the input parameters in a settings file.

\subsubsection*{Segmentation and Labeling}
Image-based operations are conducted within Napari, aiming to facilitate rapid threshold segmentation and labeling of regions. Smaller labels may be removed, and artifacts can be processed through erosion and dilation functions. Advanced functionalities, such as noise removal filters, can be integrated from the Napari library.

\subsubsection*{Property Extraction}
Based on the grayscale 3D image and the binary mask of the regions, the histograms and properties of each region can be extracted. The properties are described in Annex \ref{annex:property_extraction}. By default, four histograms are exported for each region and saved as h5ad files to disk. Properties for each label are stored in a CSV file. Whenever feasible, tasks such as Feret calculations and histogram extraction are parallelized to minimize computational time.

\subsubsection*{Exploratory Histogram Analysis}
Upon initiating the Data Viewer app (details in Annex \ref{annex:data_viewer}), histograms are resampled to an 8-bit scale (256 gray-values), and a sub-dataset comprising three randomly selected regions is generated and displayed in the "Histograms Overview". Figure \ref{fig:data_viewer} a depicts the view. Sidebar controls allow the user to modify the number of bins and regions, thereby enabling the examination of histograms from a larger number of regions (e.g., 50–100). This approach provides an initial understanding of the number of distinguishable classes and types of microstructures. However, increasing the number of regions may negatively impact the responsiveness of plot rendering. The interpretation of histograms from single regions has been extensively described in \cite{godinho_mounted_2021, godinho_3d_2023, gupta_standardized_2024}. Histogram peaks, particularly those representing dominant phases, can be enhanced using the Savitzky–Golay filter \cite{savitzky_smoothing_1964} for noise reduction in the data. Users can select up to six regions (A–F) from their label in a random subset, interactively through the heatmap or by clicking on the region in the Napari viewer. The labels selected in streamlit trigger the Napari viewer to show the center of that region. Furthermore, labels can also be selected in Napari and subsequently imported into the Data Viewer. Additionally, a “Region X” can be utilized to add a specific region of interest to the random dataset.

\subsubsection*{Peak Detection and Class Assignment}
The "Peak Finder" tab is utilized to identify the maximum gray-values for up to five classes. The input parameters of the peak finder function \cite{virtanen_scipy_2020} must be adjusted using the sliders located in the sidebar. Appropriate values should isolate only the apex of the prominent peaks that denote a class, avoiding the smaller peaks corresponding to partial volume at interphases. In figure \ref{fig:data_viewer} b an exemplary view is presented. The "Quantify all" button generates an overview of all peaks by utilizing all regions for the specified set of inputs. Brighter dots indicate an accumulation of peaks within narrow gray ranges, which can be used as a guide to select the input thresholds for each class in the input table. In the adjacent view, the number of peaks attributed to each class for the selected inputs is presented.

\subsubsection*{Volume and Mass Quantification}
The bulk and surface volume (or mass) percentages of the classes within the selected regions (labeled as region A-F) are represented as pie charts (see figure \ref{fig:data_viewer} e). If the density of the classes is specified in the input table, the volume is converted into mass percentages. The input thresholds and densities can be saved and loaded into future projects via the “Drag and drop” widget. The effectiveness of all input parameters should be visually assessed for various regions by utilizing the randomization functionality. Once the results appear reasonable, the button “Quantify all” applies the workflow to all regions. 

\subsubsection*{Global Quantification and Summary}
Once the workflow is applied to all regions, an overall statistic in the “Quantification” tab is presented, which displays the cumulative composition of all regions.  An example is depicted in figure \ref{fig:data_viewer} d. Statistics such as the number of classes per region, the number of regions, and the volume fraction analyzed can provide insight into the effectiveness of the method and the inputs (see Annex \ref{annex:data_viewer}).

\subsubsection*{Properties}
\begin{figure*}[htb!]
    \centering
    \includegraphics[width=\textwidth]{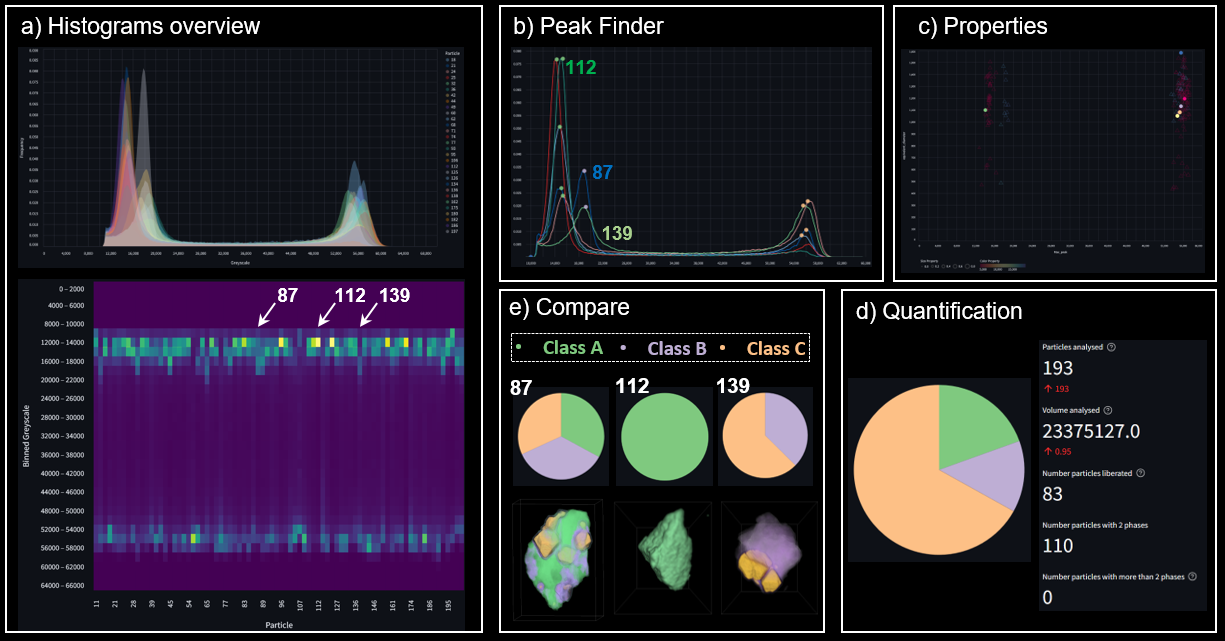}
    \caption{Figurative example of the data viewer a) Overview tab showing a subset of 50 random particles (regions). b) The peak finder plot after the parameters were optimized. Visualization of the equivalent diameter as a function of the peak gray-values. The highlighted dots correspond to the 6 regions of interest. e) Comparison between the result of the quantification (pie-plot) with the 3 microstructures of the particles. Same colors correspond to the same class (green-quartz, purple-aluminium silicates, orange-chromite)  d) Overall volume quantification and statistics. The interactivity and workflow is demonstrated in the video Annex.}
    \label{fig:data_viewer}
\end{figure*}

The "Properties" tab is utilized to compare the characteristics of all regions with the regions A-F (highlighted points, exemplarily depicted in \ref{fig:data_viewer} c). Upon pressing the "Quantify all" button, the gray-values of the peaks also become available as an analytical property. Currently, this tab serves an informative purpose and aids in the evaluation of results. For instance, plotting the peak values as a function of the equivalent diameter can reveal whether the grayscale is influenced by the size of the regions, which would require adjusting the threshold values for each class \cite{gupta_standardized_2024, godinho_3d_2023}.

\subsubsection*{Scalability and Resource Management}
Bearing in mind that 3D images can be several gigabytes in size and may contain tens of thousands of regions, the available computational resources are prone to being overwhelmed. To enhance usability, the data load can be reduced at various stages: a) the initial grayscale image may be cropped prior to segmentation; b) the export properties function can be limited to a vertical selection of the sample by specifying the start and end slices; c) interactions within the DataViewer can be performed on random subdatasets, typically comprising fewer than 100 regions, with updates occurring in near real-time. Often, only a few dozen particles are necessary to interactively determine the input parameters. Once the input parameters are saved in the project settings file, the entire workflow can be executed on the full dataset without further interaction with the viewers, which is facilitated by a Snakemake workflow (see Annex \ref{annex:scalability}).

\section*{Example}
An example of the functionalities of the software is presented for a particulate material derived from chromite ore . Details regarding the sample and its characterization have been published elsewhere \cite{ricardo_assuncao_godinho_quantitative_2023}; thus, the focus here is to describe how the characterization is facilitated by the software. The particles consist of light minerals (quartz and aluminium silicates) that exhibit similar attenuation coefficients (proportional to the gray-values), alongside chromite, which possesses a significantly higher attenuation. The software enabled the differentiation of the two light mineral classes, a task that was unattainable using the previous labor intensive 3D image processing workflow \cite{ricardo_assuncao_godinho_quantitative_2023}. It is noteworthy that identifying the appropriate parameters was accomplished in just a few hours, without the need to access the code, in contrast to the several days required in the original study. For more details we refer to Annex \ref{annex:example}. 

\section*{Impact}
The interactive tools provided by the software enable effective implementation of an analysis workflow for advanced quantitative analysis of 3D images that can be divided into less complex regions. The software offers control over the various steps of the workflow and establishes a visual link between the image and the properties of a region. Additionally, it allows users to test input parameters and assess their effects on the quantification results of selected regions in near real-time. The result is a software that is anticipated to improve: 1) the speed of the analysis using the MSPaCMAn workflow \cite{godinho_mounted_2021} to recycling and mining materials, 2) the generalization of the MSPaCMAn workflow to applications that are not necessarily related to particles, 3) accessibility for non-experts in image processing to incorporate partial volume corrections and peak-based classification in quantitative analysis, 4) Standardization of 3D image analysis.
The implementation of the MSPaCMAn workflow has been proven advantageous for various applications in the characterization of materials for the mining and recycling industries \cite{ricardo_assuncao_godinho_quantitative_2023, gupta_3d_2025, boelens_workflow_2025}. However, its application has been largely restricted to academic studies due to the required advanced knowledge of image processing, lengthy analysis times, and a comprehensive understanding of Python algorithms. The proposed software mitigates these challenges, thereby enabling the full potential of MSPaCMAn to be realized. The essential user inputs require only familiarity with the sample and do not necessitate experience in image processing or coding.
Before the introduction of a graphical user interface, users had to manually connect imaging data, histograms, material properties, and the resulting composition. This process was time-consuming and required extensive manual work, often utilizing multiple software tools such as Avizo, Orange, and Python, which demanded specialized knowledge. As a result, analyses of one sample could take weeks to complete. The graphical user interface simplifies this workflow by integrating all components into a single interface, enhancing data interaction, visualization, and analysis intuitiveness. It eliminates the need for expertise in multiple software tools and significantly reduces processing time.
Although the traditional applications of MSPaCMAn focus on particulate materials, a similar workflow could be applied to the 3D characterization of other sample types, provided the image can be segmented into discrete regions. Similarly, the reduced complexity of each region relative to the overall sample composition allows the specification of classification and quantification criteria that can enhance the detectability of phases by identifying materials using the gray-values at histogram peaks, ultimately reducing quantification uncertainty as the partial volume at interphases is accounted for. Nevertheless, the unexplored potential of the workflow is anticipated to extend across scientific fields since it addresses common imaging challenges, such as accurate classification and quantification.
This is the first open-source software providing dedicated solutions to account for partial volume and interphase blurriness, as well as histogram-based classification using peak gray-values as opposed to thresholding individual classes. Consequently, ari3d could provide an opportunity and incentive to harmonize analysis procedures across scientific fields. The initial software can be branched out to target different applications relatively easily while preserving the overall framework and fundamental equations for quantification. Modular components can be upgraded due to the modularity of the software, allowing for improvements without necessitating a complete structural change or a full re-installation. Thus, new widgets for specific actions can be incorporated into the Streamlit application, or new properties may be extracted for each region. Furthermore, image processing options can be directly integrated into the 3D viewer using Napari.

\section*{Acknowledgements}
We thank Karol Gotkowski for his help including particleSeg3D \cite{gotkowski_particleseg3d_2024} into our work. We thank Shuvam Gupta for giving valuable feedback and pointing to his code repository \href{https://github.com/ShuvamGupta/mspacman}{https://github.com/ShuvamGupta/mspacman}.
This work has received financing from the Deutsche Forschungsgemeinschaft within SPP 2315, project \href{https://www.sciencedirect.com/science/article/pii/S2949673X25000038\#spnsr1}{470202518}. Part of this work was funded by HELMHOLTZ IMAGING, a platform of the Helmholtz Information \& Data Science Incubator. Furthermore, the project benefited from the Deutsche Forschungsgemeinschaft (DFG, German Research Foundation) – Project-ID 414984028 – CRC
1404 FONDA

\section*{Competing Interest}
The authors declare no competing interests.
\\

\begin{contributions}
\textbf{Jan Philipp Albrecht} Conceptualization, Data Curation, Formal Analysis, Investigation, Methodology, Project Administration, Software, Validation, Visualization, Writing – Original Draft Preparation, Writing – Review \& Editing
\textbf{Deborah Schmidt} Funding Acquisition, Investigation, Project Administration, Resources, Supervision, Writing – Review \& Editing
\textbf{Christina Hübers} Software
\textbf{Jose Ricardo Assunção Godinho} Conceptualization, Data Curation, Formal Analysis, Funding Acquisition, Investigation, Methodology, Project Administration, Resources, Software, Supervision, Validation, Visualization, Writing – Original Draft Preparation, Writing – Review \& Editing


\end{contributions}

\section*{Declaration of generative AI and AI-assisted technologies in the writing process}

During the preparation of this work the author(s) used ChatGPT 4o in order to reword the text. After using this tool/service, the author(s) reviewed and edited the content as needed and take(s) full responsibility for the content of the publication.

\section*{References}
\addcontentsline{toc}{section}{References}
\bibliography{references}
\clearpage
\newpage



\onecolumn
\appendix



\setcounter{table}{0}
\renewcommand{\thetable}{S\arabic{table}}
\renewcommand{\tablename}{Supplementary Table} 

\setcounter{figure}{0}
\renewcommand{\thefigure}{S\arabic{figure}}
\renewcommand{\figurename}{Supplementary Figure} 

\section{Workflow \& Software Components}
\label{annex:workflow}

This section provides a detailed overview of the software-supported workflow designed for material classification and quantification. The workflow relies on the reduced complexity of each region, relative to the overall composition of the sample and enables the establishment of classification and quantification criteria. The key objectives are: 
\begin{itemize}
    \item Enhance the detectability of phases by identifying the materials based on the gray-values at histogram peaks, which are generally better resolved at the regional level.
    \item Minimize quantification uncertainty by accounting for the partial volume effects at interphases. 
\end{itemize}

The workflow encompasses five fundamental steps:
\begin{enumerate}
    \item General assessment of the 3D image. The purpose of this step is to determine the suitability of the method in contrast with traditional alternatives. In principle, this method becomes advantageous if the objects to be quantified are relatively small so that the voxels representing interphases are a significant fraction of the objects. In these cases, it is beneficial to classify the image using peak gray-values instead of gray-value ranges; and a necessity to account for the partial volume at the interphases.
   
    \item Segmentation and labeling of the regions containing the materials to be classified and quantified. This step can be implemented using any traditional segmentation method, e.g. thresholding and AI-based. Pre-processing of the grayscale image could improve the segmentation results, e.g. by applying a noise removal filter. It may also be beneficial to process the result of the segmentation to remove artifacts, e.g. removing small regions or morphological operations.
    
    \item Extraction of the region's properties. For each region, a set of geometric properties and the histograms are extracted, thus reducing the image into lists of numerical data, “properties” and “histograms”, whereby each row corresponds to a region. The histogram of the region’s surface is extracted to get the surface composition as well as to better quantify the partial volume between the region and the sample’s matrix.
    
    \item Definition of the criteria that identify the various classes. A class is defined as a material or a group of materials that can be identified by a specific property or combination of properties. The properties are obtained from the histogram and the properties files. Therefore, this step requires some preliminary knowledge about the sample and the interpretation by a user of the properties and histograms data extracted from step 3. The most effective way of doing this is by comparing the image of the region with the histogram and properties. The most common feature to identify a class is the gray-value of the histogram peaks.   
    
    \item Quantification of each class. By applying the criteria defined in step 4, a list of classes per region is created. Knowing the classes allows the calculation of the partial volume at interphases between the classes inside each region.
\end{enumerate}

\begin{figure*}[htb!]
    \centering
    \includegraphics[width=0.8\textwidth]{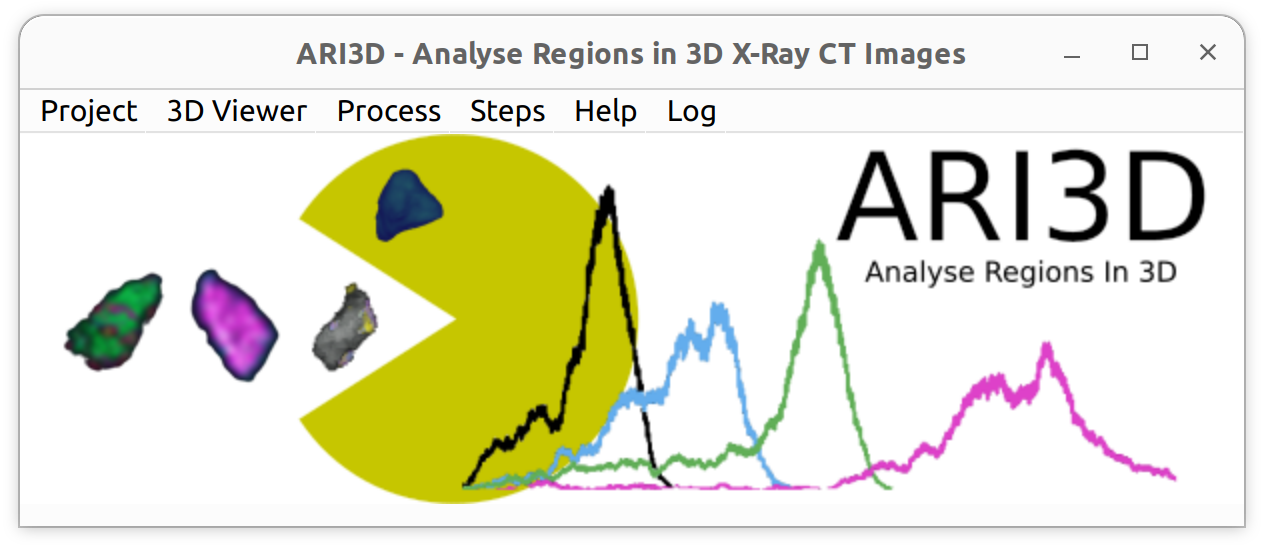}
    \caption{View on the Main Menu of the Software. This serves as control window for the entire workflow.}
    \label{fig:main_menue}
\end{figure*}

A visual representation of this workflow is provided in Supplementary Figure \ref{fig:schamtic_view}. The graphical user interface (see Figures~\ref{fig:main_menue}, \ref{fig:step_3_popup}, and 
\ref{fig:data_viewer}) supports each step, offering intuitive access to tools for visualization, interaction, and quantification. 

It is important to note that the scientific rationale underpinning each step has been validated in numerous studies; therefore, the focus of this article is primarily on the associated software.

\begin{figure*}[htb!]
    \centering
    \includegraphics[width=0.8\textwidth]{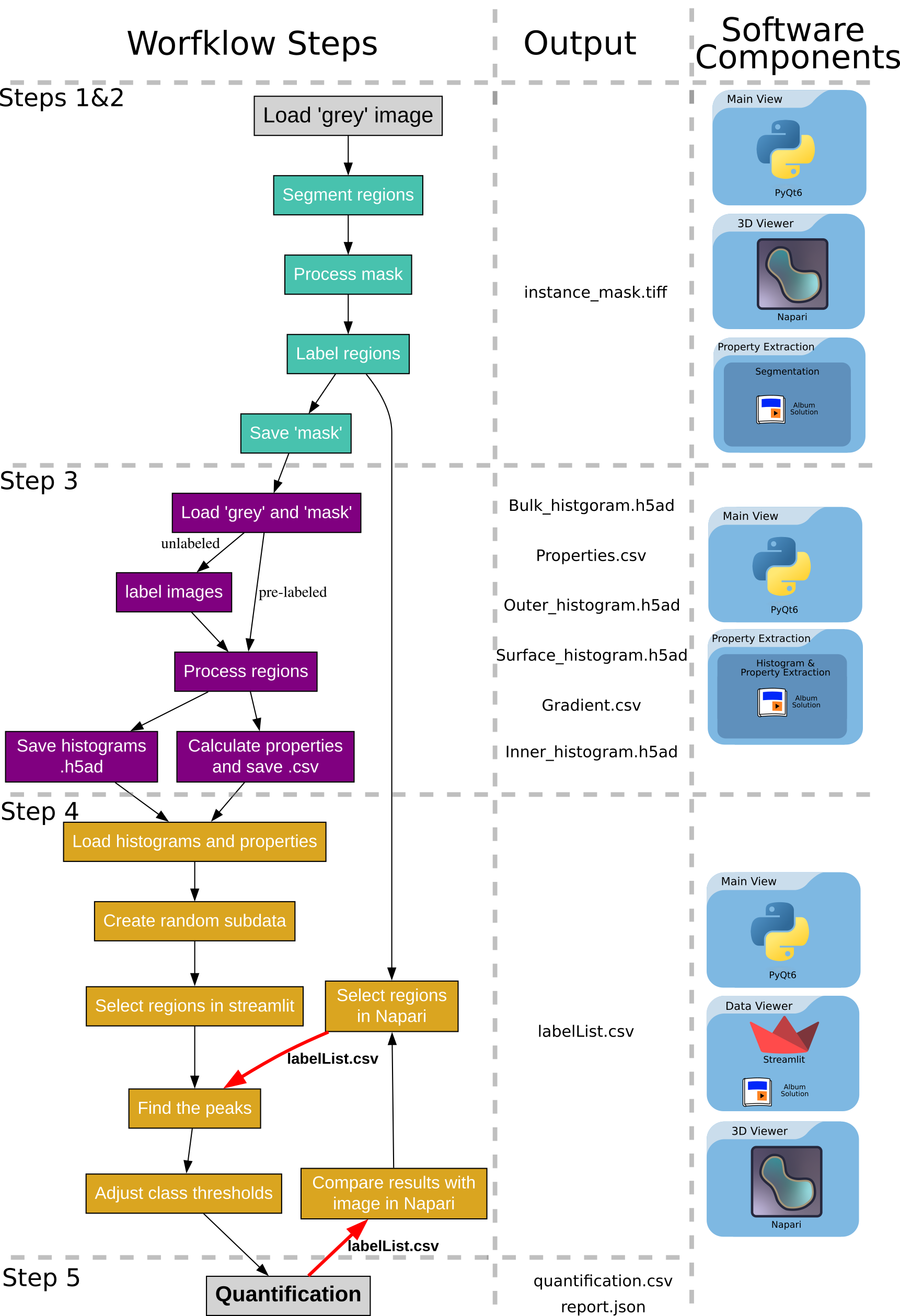}
    \caption{Scheme of the main functionalities of the software (as workflow steps) that link to the different components of the software required for the workflow execution. Additionally the column \textit{Output} lists the respective files created.  Note that all workflow components can be steered via the Main View.}
    \label{fig:workflow}
\end{figure*}

\clearpage
\newpage

\section{User parameters}
\label{annex:user_parameters}

This section provides an overview of the input parameters required for executing the complete workflow. These parameters are defined in the \texttt{parameters.yml} file and can be adjusted based on sample characteristics and analysis goals.

The parameters are grouped into four categories:
\begin{itemize}
    \item \textbf{Mineralogy Analysis}
    \item \textbf{Property Extraction}
    \item \textbf{Segmentation}
    \item \textbf{Manual Segmentation}
\end{itemize}

Supplementary Table~\ref{tab:parameters} summarizes all available parameters, along with their descriptions and functional relevance.

\begin{table}[h!]
\centering
\begin{adjustbox}{}
\resizebox{0.7\textwidth}{!}{%
\begin{tabular}{l l p{7cm}}
\hline
\textbf{Category} & \textbf{Name} & \textbf{Description} \\
\hline
Mineralogy Analysis & DensityA & Density of Material Class A \\
Mineralogy Analysis & DensityB & Density of Material Class B \\
Mineralogy Analysis & DensityC & Density of Material Class C \\
Mineralogy Analysis & DensityD & Density of Material Class D \\
Mineralogy Analysis & DensityE & Density of Material Class E \\
Mineralogy Analysis & MaxGreyValueA & Maximal gray-value threshold for Phase 1 \\
Mineralogy Analysis & MaxGreyValueB & Maximal gray-value threshold for Phase 2 \\
Mineralogy Analysis & MaxGreyValueC & Maximal gray-value threshold for Phase 3 \\
Mineralogy Analysis & MaxGreyValueD & Maximal gray-value threshold for Phase 4 \\
Mineralogy Analysis & MaxGreyValueE & Maximal gray-value threshold for Phase 5 \\
Mineralogy Analysis & background\_q & Background gray-value for quantification \\
Mineralogy Analysis & binInput & Absolute number of bins used during analysis \\
Mineralogy Analysis & enableSavgol & Enable Savitzky-Golay smoothing filter (boolean) \\
Mineralogy Analysis & enablePVB & Enable Partial Volume Blurr Correction (boolean) \\

Mineralogy Analysis & gray\_value\_width & Width between valleys on either side of a peak \\
Mineralogy Analysis & horizDistance & Minimum horizontal distance between peaks \\
Mineralogy Analysis & min\_frequency & MinimumConceptualization height of peaks from bottom \\
Mineralogy Analysis & prominence & Minimum prominence of a peak \\
Mineralogy Analysis & savgolInput & Intensity of Savitzky-Golay smoothing filter \\
Mineralogy Analysis & vertDistance & Minimum vertical distance between peaks \\
Property Extraction & aspect\_ratio & Calculate aspect ratio of segmented objects (boolean) \\
Property Extraction & background & Background gray-value threshold for property extraction \\
Property Extraction & basic\_properties & Extract basic properties (boolean) \\
Property Extraction & end\_slice & Last slice to process (-1 for all) \\
Property Extraction & entropy & Calculate entropy of the image (boolean) \\
Property Extraction & euler & Calculate Euler number (boolean) \\
Property Extraction & feret\_angle & Angle increment for calculating Feret diameter \\
Property Extraction & ferets & Extract Feret diameters (boolean) \\
Property Extraction & histogram & Generate histogram of gray-values (boolean) \\
Property Extraction & inertia & Calculate inertia tensor properties (boolean) \\
Property Extraction & mean\_max & Compute mean and max gray-values (boolean) \\
Property Extraction & mesh\_spacing & Spacing of mesh points for 3D reconstruction \\
Property Extraction & moments & Compute statistical moments (boolean) \\
Property Extraction & num\_threads & Number of CPU threads to use (-1 for all available) \\
Property Extraction & save\_labels & Boolean flag to save labels of segmented regions \\
Property Extraction & solidity & Compute solidity of objects (boolean) \\
Property Extraction & start\_slice & First slice to process (-1 for all) \\
Property Extraction & voxel\_size & Voxel resolution of the image in micrometers \\
Segmentation & apply\_to\_slice & Apply segmentation to a specific slice (-1 for all) \\
Segmentation & average\_particle\_size\_mm & Average particle size in millimeters (used for segmentation) \\
Segmentation & model\_path & Path to deep learning segmentation model \\
Segmentation & remove\_small & Minimum size of objects to retain in segmentation \\
Segmentation & save\_seg\_labels & Save segmented labels (boolean) \\
Segmentation & voxel\_size\_mm & Voxel size in millimeters \\
Manual Segmentation & dilation\_erosion\_operations & Sequence of dilation (1) and erosion (2) operations \\
Manual Segmentation & manual\_thresholding & Threshold value for manual segmentation \\
\hline
\end{tabular}
}
\end{adjustbox}
\caption{Input parameters categorized by workflow function.}
\label{tab:parameters}
\end{table}

\clearpage
\newpage

\section{Code Structure}
\label{annex:code_structure}

This reference aims to assist researchers and developers in further developing and maintaining the software efficiently. By adhering to this structured approach, we ensure modularity, maintainability, and ease of extension for future enhancements.

Our source code is structured to optimize \textbf{reusability} and \textbf{separation of responsibilities}, ensuring that each component effectively represents a specific part of the application. The structure is as follows:
\begin{samepage}
\begin{verbatim}
-ari3d
    |--solutions [...] 
    |-- src
    |   \--ari3d
    |       |-- gui
    |       |	|-- controller [...]
    |       |	|-- model [...]
    |       |	|-- view [...]
    |       |	|-- ari3d.py
    |       |	\-- ari3d_logging.py
    |       |-- resources
    |       |	|-- images [...]
    |       |	|--models [...]
    |       |	|-- default_values.py
    |       |	\-- parameters.yml
    |       |-- solutions
    |       |	|-- data_viewer [...]
    |       |	|-- property_extraction [...]
    |       |	\-- segmentation [...]
    |       |-- utils [...]
    |       \-- workflow
    |           |-- ari3d_cli.py
    |           \-- Snakefile
    |-- album_catalog_index.db
    |-- album_catalog_index.json
    |-- LICENSE
    |-- MANIFEST.in
    |-- README.md
    |-- pyproject.toml
    \-- solution.py
\end{verbatim}
 
Items marked with \texttt{[...]} denote substructures that are not illustrated here for the sake of brevity.

\end{samepage}

\subsection*{Root-Level Components}
At the root level, we include folders and files that are essential for packaging, distribution, documentation, and licensing:

\begin{itemize}
    \item \textbf{Packaging \& Distribution}: We utilize setuptools for packaging. This requires \textit{pyproject.toml}, and \textit{MANIFEST.in}. Readers interested in packaging details can refer to the official setuptools documentation. We further provide a installation of our snakemake workflow as an album solution (\textit{solution.py}).
    
    \item \textbf{Documentation \& Usage}: The \textit{README.md} file provides installation and maintenance instructions. It is displayed on the GitLab project homepage.
    Licensing: The \textit{LICENSE} file ensures compliance with legal requirements. We have chosen the GPLv3 license to align with PyQt6 licensing requirements.
    
    \item \textbf{Workflow \& Solution Updates}: \textit{album\_catalog\_index.db} and \textit{album\_catalog\_index.json}, along with the solutions folder, support workflow step updates through album solutions. These files and folders should not be manually modified. Instead, any updates should follow the instructions outlined in \textit{README.md}.
\end{itemize}

\subsection*{Source Code Structure (src)}
The core application logic is located in the src directory. It is further divided into five key components:
\begin{itemize}
    \item \textbf{GUI (Graphical User Interface)}: This folder encompasses the user interface logic and design. The organization follows the Model-View-Controller (MVC) principle, a well-established software engineering paradigm for UI development:
    \begin{itemize}
        \item view/ – Design files for each window.
        \item controller/ – Logic applied to each UI element.
        \item model/ – Helper classes that support the controller’s functionality.
        \item ari3d.py – Main entry point controller element.
        \item ari3d\_logging.py – Logging functionality for the entire code logic.
    \end{itemize}    
    
    \item \textbf{resources (Non-Code Assets)}: This folder contains all essential assets that, while not directly involved in logic execution, are necessary for overall functionality:
    \begin{itemize}
        \item images/ – UI icons and graphical assets.
        \item models/ – Pre-trained deep learning models for segmentation. Will be downloaded once when needed. They are not shipped during installation.
        \item default\_values.py – Contains predefined software settings.
        \item parameters.yml – A configuration file listing workflow parameters and their default values, which users can modify. Please look below for further information.
    \end{itemize}
    
    \item \textbf{solutions (Modular Workflow Steps)}: This directory contains the implementation of the three primary workflow steps. These steps are modularized using album. Instructions for updating these steps can be found in the project documentation.
    \begin{itemize}
        \item segmentation/ – Segmentation via ParticleSeg.
        \item property\_extraction/ – Properties and histogram extraction.
        \item data\_viewer/ – Data visualization and quantification.
    \end{itemize}

    \item \textbf{utils (Helper Functions)}: This folder contains utility functions for managing file folder, and image operations utilized throughout the codebase.
    
    \item \textbf{workflow (Non-GUI Execution)}: Contains the files required to execute the workflow in a non-graphical environment utilizing Snakemake:
    \begin{itemize}
        \item \textit{ari3d\_cli.py} – Entry point to the Snakemake workflow execution.
        \item \textit{Snakefile} – Defines the workflow steps in Snakemake.
    \end{itemize}
\end{itemize}

\subsection*{Implementation Details}
\label{section:imp_detail}

We present the links to the implementations utilized in our software in Supplementary Table \ref{tab:imp-detail}. Furthermore, we would like to direct attention to \href{https://github.com/ShuvamGupta/mspacman}{https://github.com/ShuvamGupta/mspacman}, which details the implementation behind \cite{gupta_3d_2025} that was used in this work.

\begin{table}[H]
\centering
    \begin{tabular}{lll}
    \hline
    \textbf{Function} & \textbf{Import structure}  & \textbf{Package Name}\\
    \hline
    Savitzky-Golay Filter & \href{https://docs.scipy.org/doc/scipy/reference/generated/scipy.signal.savgol_filter.html}{\texttt{scipy.signal.savgol\_filter}} & scipy \\
    find peaks & \href{https://docs.scipy.org/doc/scipy/reference/generated/scipy.signal.find_peaks.html}{\texttt{scipy.signal.find\_peaks}} & scipy  \\
    binary\_erosion & \href{https://docs.scipy.org/doc/scipy/reference/generated/scipy.ndimage.binary_erosion.html}{\texttt{scipy.ndimage.binary\_erosion}} & scipy \\
    regionprops & \href{https://scikit-image.org/docs/0.25.x/api/skimage.measure.html#skimage.measure.regionprops}{\texttt{skimage.measure.regionprops}} & scikit-image \\
    remove small objects & \href{https://scikit-image.org/docs/0.25.x/api/skimage.morphology.html#skimage.morphology.remove_small_objects}{\texttt{skimage.morphology.remove\_small\_objects}} & scikit-image \\
    marching cubes & \href{https://scikit-image.org/docs/0.25.x/auto_examples/edges/plot_marching_cubes.html}{\texttt{skimage.measure.marching\_cubes}} & scikit-image \\
    mesh surface area & \href{https://scikit-image.org/docs/0.25.x/api/skimage.measure.html#skimage.measure.mesh_surface_area}{
    \texttt{skimage.measure.mesh\_surface\_area}} & scikit-image \\
    
    \hline
    \end{tabular}
\caption{Links to relevant processing functions.}
\label{tab:imp-detail}

\end{table}

\subsection*{Modularity}
\label{section:modularity}

We utilize album to modularize workflow steps, ensuring they are exchangeable and updatable. Album executes each workflow step (solutions - in this context) in its own environment. Hence, compatibility issues between software and solutions are mitigated. Furthermore, album enables the distribution of these solutions in a decentralized manner, referred to as a catalog. This approach allows for the updating of individual steps without the necessity of altering the software, as steps can be reinstalled at a later time, for example after an update.  For illustration, we describe the tasks necessary to update the property extraction process from both developer and user perspectives. First, the developer updates the solution \textit{property\_extraction} in the \textit{src} directory of the Git project. For example, by adding a property that can be extracted from a region. Second, the developer uses album to deploy the solution to the repository, which serves as both the catalog (distribution source) and codebase. Third, the user utilizes the graphical user interface of our software to check for updates to any steps. By confirming the update process, the corresponding new solution is installed as the final step on the user’s machine.

\clearpage
\newpage

\section{Project Structure}
\label{annex:project_structure}

The first step in analyzing a new sample is to create a project, which can be initiated via the menu: \textbf{Project → Create Project}. The user selects a folder containing \textit{TIFF} images as input data. Upon creation, the software generates a 3D volume in ZARR format from the input images and organizes the project structure within the same folder.  Below is an example of the resulting directory structure for a completed project:

\begin{samepage}
\begin{verbatim}
- MyProject
        |-- analysis
        |	|-- Bulk_histogram.h5ad
        |	|-- Gradient.csv
        | 	|-- Inner_histogram.h5ad
        | 	|-- labelList.csv
        | 	|-- Outer_histogram.h5ad
        | 	|-- Properties.csv
        | 	\-- Surface_histgoram.h5ad
        |-- gray
        |	\-- [*].tiff
        |-- mask
        |	\-- instance_mask.tiff
        |-- project
        |	|- MyProject.zarr
        |	\--[...]
        \-- report
            |- parameters.yml
            |- quantification.csv
            \-- report.json
\end{verbatim}

\begin{itemize}
    \item \textbf{[*]} indicates a wildcard for multiple TIFF image files.
    \item \textbf{[…]} denotes additional folders omitted for clarity, but necessary for workflow execution.
\end{itemize}
\end{samepage}

Once the project is initialized, it can be loaded at any time via \textbf{Project → Load Project}. The workflow does not need to be completed in one session — the software supports pausing and resuming from any step. This standardized folder structure promotes both reproducibility and flexibility throughout the analysis process. An overview of the essential files and their role in the workflow is presented in Supplementary Table~\ref{tab:project_files}.

\begin{table}[ht]
\centering
\begin{tabular}{l p{7cm} l}
\hline
\textbf{File name or pattern} & \textbf{Description} & \textbf{Workflow Step} \\
\hline
Bulk\_histgoram.h5ad & Unbinned histogram of the volume for each particle. & Step 2 \\
Outer\_histogram.h5ad & Unbinned histogram of the outer volume of each particle. & Step 2 \\
Surface\_histogram.h5ad & Unbinned histogram of the 1-voxel depth surface of each particle. & Step 2 \\
Inner\_histogram.h5ad & Unbinned histogram of the inner volume of each particle. & Step 2 \\
Gradient.csv & Intensity surface volume gradients. & Step 2 \\
LabelList.csv & List of particle properties, identified by their labels, for highlighting. & Step 1,2,3 \\
Properties.csv & Detailed particle properties. & Step 1,2,3 \\
instance\_mask.tiff & Instance segmentation mask of the entire volume. & Step 1 \\
MyProject.zarr & ZARR container for input files. & Step 1 \\
parameters.yml & Workflow execution parameters. & All steps \\
quantification.csv & Volume mineral class quantification per particle. & Step 5 \\
report.json & Quantification report including summary statistics. & Step 5 \\
\hline
\end{tabular}
\caption{Summary of files, descriptions, and the workflow steps they relate to.}
\label{tab:project_files}
\end{table}

\clearpage
\newpage

\section{Property Extraction}
\label{annex:property_extraction}

\begin{figure*}[htb!]
    \centering
    \includegraphics[width=0.8\textwidth]{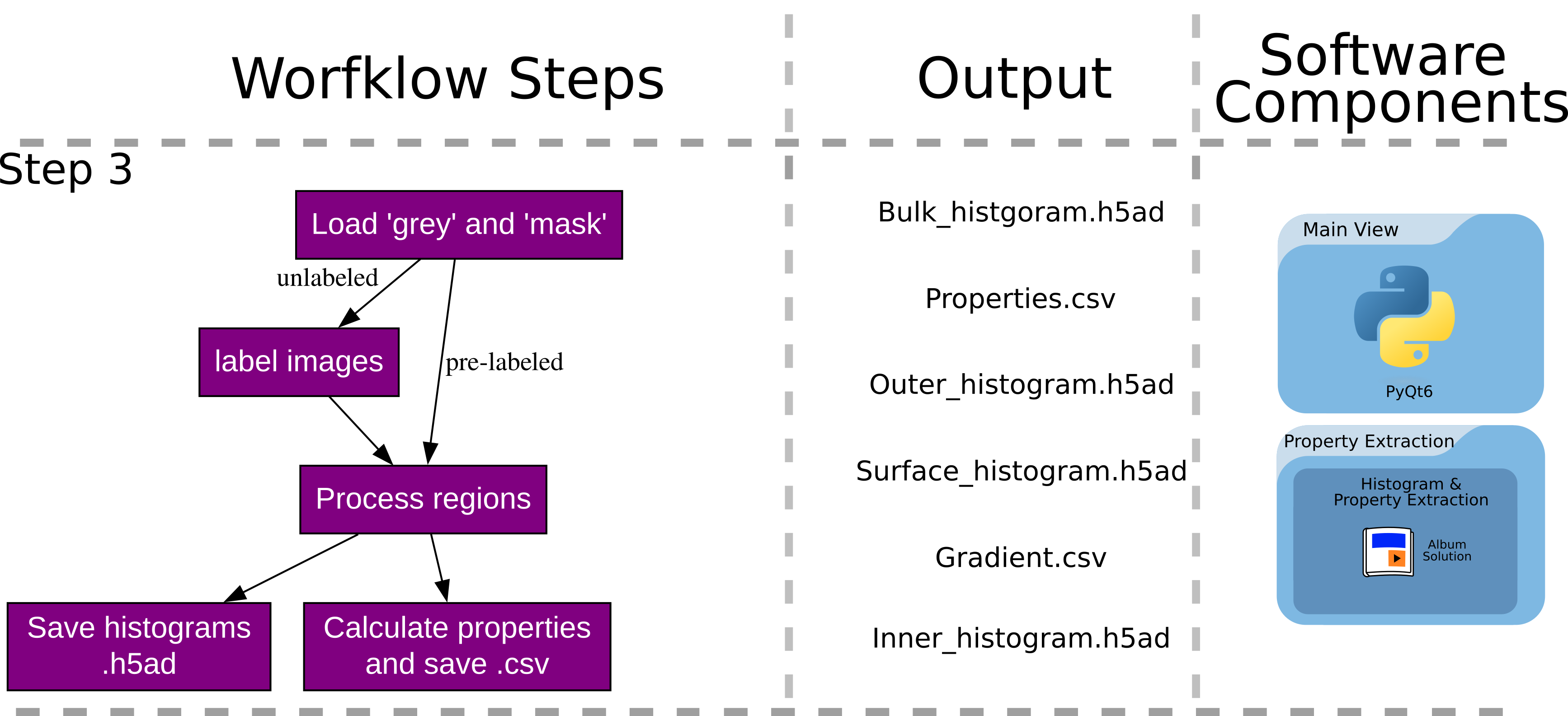}
    \caption{Overview of Step 3: loading, processing, and extracting histograms and properties from labeled 3D regions.}
    \label{fig:step_3}
\end{figure*}

Step 3 of the workflow (see Figure~\ref{fig:step_3}) focuses on extracting geometric and intensity-based properties from segmented regions. This process begins by loading the \texttt{gray} and \texttt{mask} images from their respective folders.


\subsection*{Input Processing}

\begin{itemize}
    \item If the input mask is not labeled, the software automatically labels non-touching regions.
    \item If the \textbf{Save labels} checkbox is selected, the labeled image is saved as a 16-bit TIFF.
    \item To manage memory usage, the user may restrict the loaded volume by specifying a \textbf{Start slice} and/or \textbf{End slice}.
    \item Regions touching the image boundaries or smaller than the user-defined threshold (\texttt{remove\_small}) are removed.
    \item A surface mesh is created for each remaining region.
\end{itemize}

\subsection*{Outputs}

\begin{figure*}[htb!]
    \centering
    \includegraphics[width=0.8\textwidth]{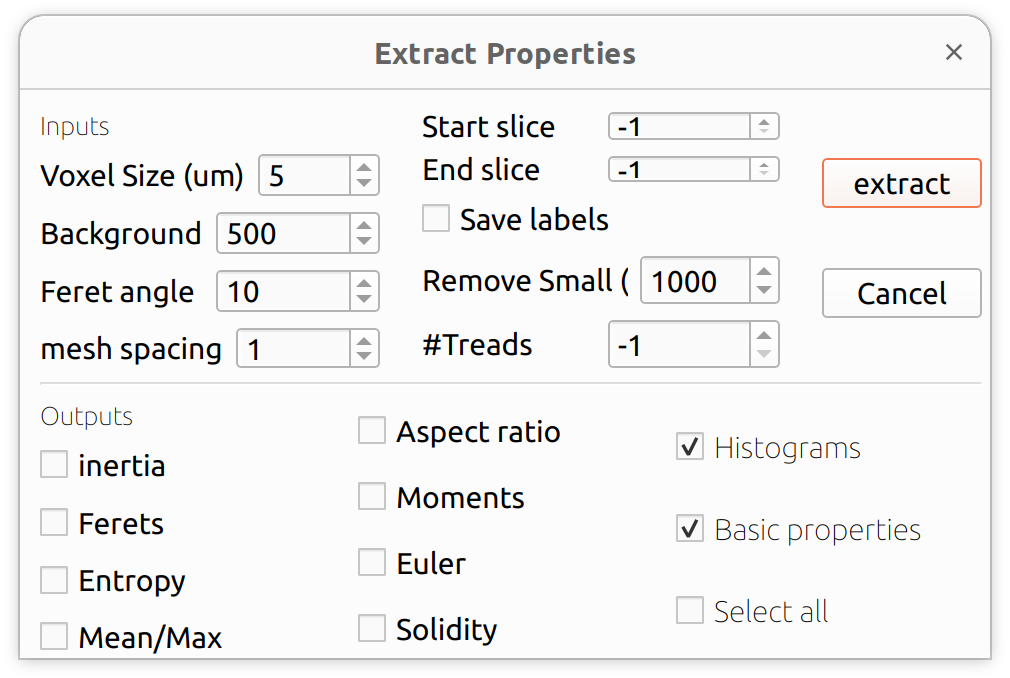}
    \caption{Popup window for choosing processing parameters and selecting output properties.}
    \label{fig:step_3_popup}
\end{figure*}

By default, the software exports both histograms and basic geometric properties. Figure~\ref{fig:step_3_popup} shows the popup window where additional processing options can be configured.

\subsubsection*{Input Parameters}

\begin{itemize}
    \item \textbf{Voxel size}: If different from 1, the outputs are exported in um.
    \item \textbf{Background}: Input the average gray-value of the sample’s matrix, i.e., where the regions are dispersed. An approximate value is sufficient.
    \item \textbf{Feret angle}: This is only used if the output “Ferets” is ticked. The smaller the angle the more realistic the measurement is, although the slower and heavier on RAM the computation. The default of 10 is usually a fair compromise.
    \item \textbf{Mesh spacing}: distance in voxels used to generate the surface mesh. 1 gives the best resolution of surface properties. If too heavy on RAM or if surface details are not crucial, higher values can be used. 
    \item \textbf{Remove small}: the minimum number of voxels required per region. Only larger regions are processed. Depending on the purpose of the study, smaller regions can and should be ignored. A typical good value is 800.  
    \item \textbf{Threads}: Number of processor threads used for the parallelized functions. The default -1 uses the full computer power. Can be reduced if the program crashes or if other tasks are running in the background. 
\end{itemize}

\subsubsection*{Output Options}

\begin{itemize}
    \item \textbf{Histograms}: 
    \begin{itemize}
        \item Bulk histograms represent all voxels in a region
        \item Surface mesh histograms represent only the voxels on the surface of a region
        \item Outer histograms represent only the voxels on outer n-layers of the particle, where n is determined from the gradient of gray-values for each region
        \item Inner histograms represent the voxels in the core of the particle (bulk – outer voxels)
    \end{itemize}
    \item \textbf{Basic Properties}:
    \begin{itemize}
        \item Surface area: sum of the area of all triangles in the surface mesh.
        \item Volume: number of voxels in a region multiplied by the cubic of the voxel size.  
        \item Minimum: minimum gray-value in a region
        \item Maximum: maximum gray-value in a region 
        \item Mean: mean gray-value of all voxels in a region
        \item Centroid: the 3D coordinates of the center of a region
    \end{itemize}
    \item \textbf{Optional Properties}:
    \begin{itemize}
        \item Ferret: Parameters that characterize the shape and size of a defined region. 
        \item Mean/Max: This value is calculated by dividing the mean gray-value by the maximum gray-value within a region. 
        \item Inertia: Measures the distribution of voxel mass around the principal axes, indicating how resistant the shape is to rotational motion.
        \item Entropy: Quantifies the complexity or randomness in the voxel intensity distribution within the 3D image.
        \item Aspect Ratio: Describes the proportional relationship between the dimensions of the bounding box enclosing the 3D object.
        \item Moments:  Statistical values that capture the spatial distribution and shape characteristics of the 3D object.
        \item Euler: Represents the topological structure by counting objects, holes, and cavities in the 3D shape.
        \item Solidity: Measures the compactness of the object by comparing its volume to the volume of its convex hull.
    \end{itemize}

\end{itemize}

\clearpage
\newpage

\section{DataViewer}
\label{annex:data_viewer}

The DataViewer module facilitates interactive data exploration and classification. It is implemented using \textbf{Streamlit}, while the 3D Viewer relies on \textbf{napari}, a Python-based framework for visualizing multi-dimensional images.

As illustrated in Figure~\ref{fig:step_4_5}, the DataViewer is integrated with the 3D Viewer, though most computational operations are handled by the Streamlit interface (see Figure~\ref{fig:DataViewer}). Several tabs and bars are available for user interaction and will be presented thereafter:

\begin{figure*}[htb!]
    \centering
    \includegraphics[width=0.8\textwidth]{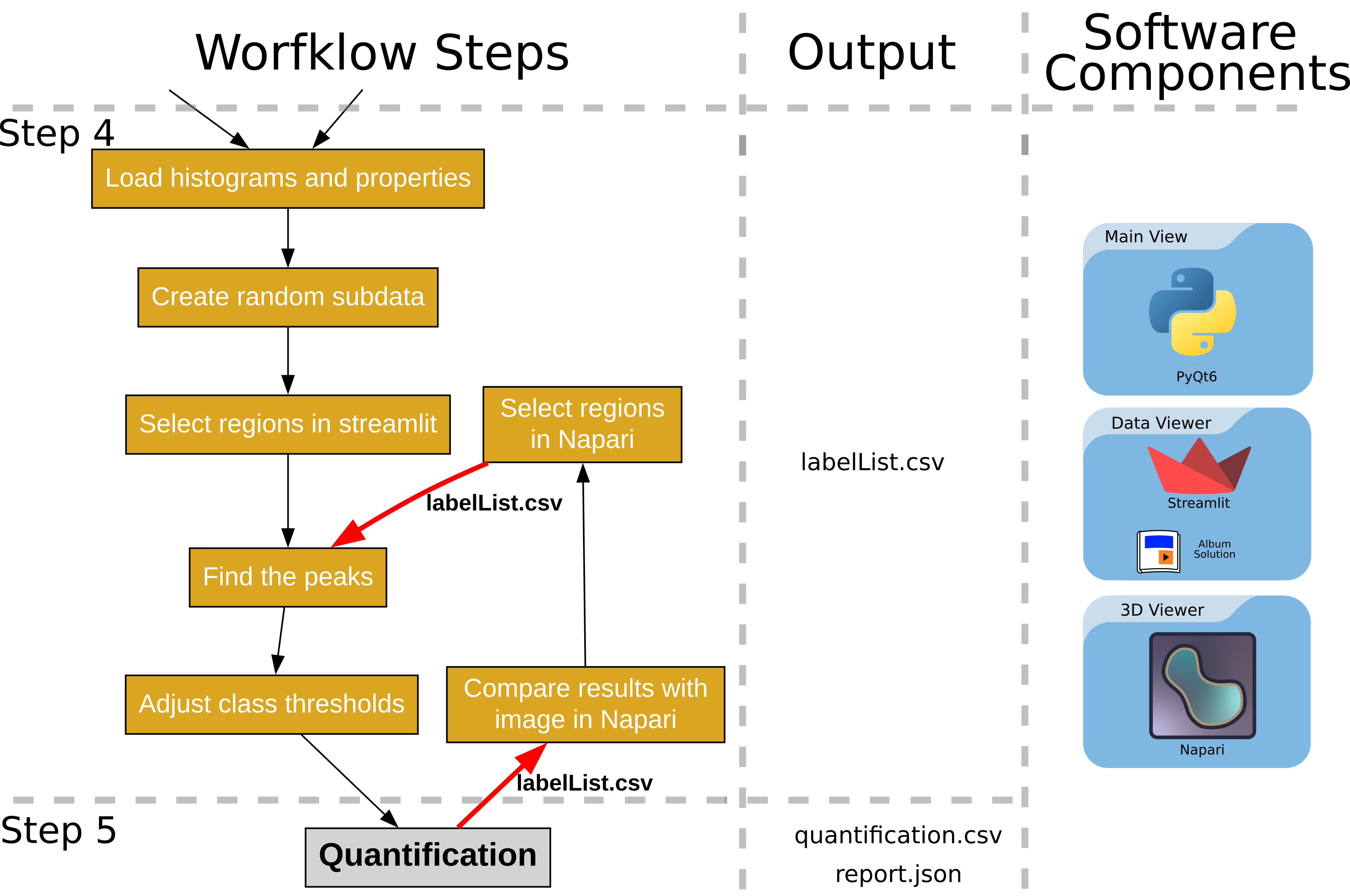}
    \caption{Step 4 and 5 of the workflow together with their software components.}
    \label{fig:step_4_5}
\end{figure*}

\begin{figure*}[htb!]
    \centering
    \includegraphics[width=0.8\textwidth]{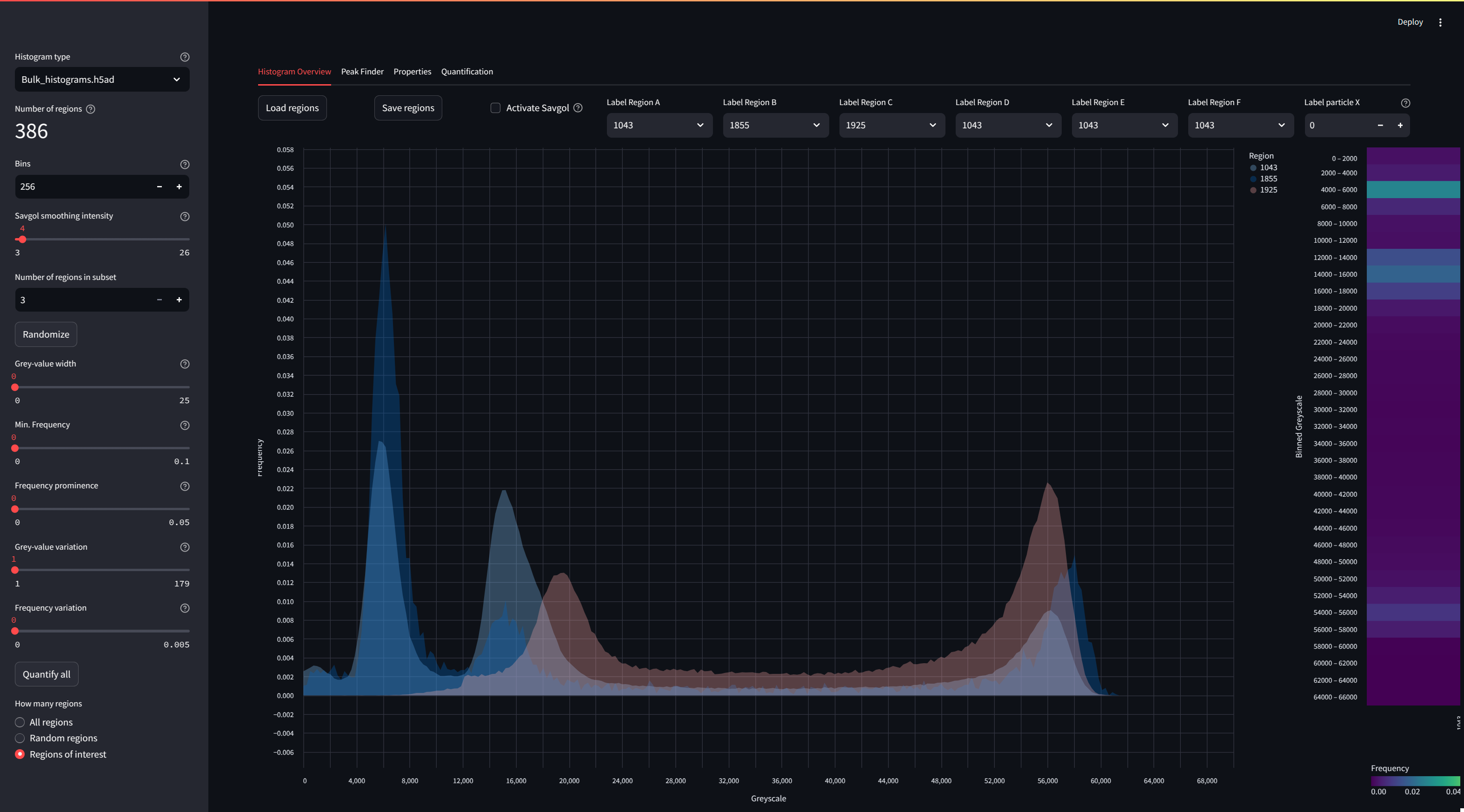}
    \caption{DataViewer web-page view. The sidebar contains inputs that automaticaly updated upon change. The tabs show the analysis results.}
    \label{fig:DataViewer}
\end{figure*}

\newpage
\subsection*{Sidebar Options}

\begin{itemize}
    \item \textbf{Histogram Type:} Select which histogram to display (e.g., bulk, surface, inner, outer). Bulk histograms are generally preferred for material classification. Choose other histograms to assess the surface composition and the amount of inner vs outer partial volume effect.

    \item \textbf{Bins:} Adjusts the number of histogram bins used in plotting. Note: The x-axis always represents the original gray-value scale (e.g., 0–65535 for 16-bit images), regardless of bin count.

    \item \textbf{Number of Regions \& Randomization:} Clicking \textit{Randomize} generates a sub-sample of the selected number of regions. This updates the label menus and plots accordingly.

    \item \textbf{Savgol Smoothing Intensity:} Controls the smoothing level for histograms. Affects both plots and the peak-finder function, but quantification is performed on raw data.

    \item \textbf{Peak Finder Sliders:}
    \begin{itemize}
        \item \textbf{gray-value Width}
        \item \textbf{Minimum Frequency}
        \item \textbf{Prominence}
        \item \textbf{Horizontal \& Vertical Distance}
        \item \textbf{gray-value/Frequency Variation}
    \end{itemize}
    These sliders configure the \texttt{find\_peaks} algorithm used to detect gray-value peaks. Should be adjusted according to the sample’s characteristics. Follow link in Implementation Detail in \ref{section:imp_detail}.

    \item \textbf{Quantify All:} Applies the classification to all regions and generates peak statistics. Useful for assessing class distributions and classification accuracy.

    \item \textbf{Region Display Options:}
    \begin{itemize}
        \item \textbf{All regions} – Display all available regions.
        \item \textbf{Random regions} – Display a random sub-sample.
        \item \textbf{Regions of interest (A–F, X)} – Display manually selected regions.
    \end{itemize}
\end{itemize}

\subsection*{Overview Tab}

\begin{itemize}
    \item \textbf{Plot Overview:} Displays histograms of selected regions. Each curve represents one region, and frequencies are normalized.

    \item \textbf{Heatmap:} Regions on the x-axis, gray-values on the y-axis, color representing frequency. Useful to identify consistent gray-value peaks across regions. Brighter colors signify higher percentages, thus are a good indication of peaks. A colorful horizontal band across multiple regions indicate a common microstructure. Note: clicking in one region or ctrl+click in multiple regions allows selecting specific histograms.

    \item \textbf{Region Dropdown Menus:} Allows manual selection of regions A–F and a specific region X. Use \textbf{Save} to write selections to \texttt{LabelList.csv} or \textbf{Load} to import from it.
    Helpful to compare regions with similar peak positions in order to help define the threshold ranges for each material. Further, helpful to select at least one region of each phase to test the peak finder parameters. Region X allows selecting a specific region even if it is not included in the random subset.
    
    \item \textbf{Savgol Box:} Toggles histogram smoothing using Savitzky-Golay.
\end{itemize}

\subsection*{Peak Finder Tab}

\begin{itemize}
    \item \textbf{Threshold Table:} Defines gray-value thresholds for material class identification. The maximum value defines the upper boundary; the lower boundary is inherited from the previous class.

    \item \textbf{Density Field:} If set to 1, quantification is in volume fraction. Other values convert volumes to mass fractions.

    \item \textbf{Background gray-value (\texttt{BackgroundT}):} Approximate background matrix value, excluding particles.

    \item \textbf{Peak Plot:} Overlays peak positions (colored dots) on line histograms. Peak classification is based on thresholds defined in the table.

    \item \textbf{Pie Charts:} Show the volume or mass distribution of each class within the region and on the region’s surface.
\end{itemize}

\subsection*{Properties Tab}

\begin{itemize}
    \item \textbf{Property Scatterplot:} Displays selected region properties (e.g., size, shape, gray-value) with customizable axes, color scale, and marker size.
    
    \item \textbf{Post-Quantification Enhancements:} After running \textit{Quantify All}, additional peak-derived properties become available for plotting.
\end{itemize}

\subsection*{Quantification Tab}

\begin{itemize}
    \item \textbf{Pie Charts:} Represent volume or mass fractions across all analyzed regions.
    \item \textbf{Region Statistics:}
    \begin{itemize}
        \item \textbf{Regions Analyzed:} Lower numbers may indicate undetectable peaks.

        \item \textbf{Volume Analyzed:} Should approach 1.0 for effective classification. Values below 0.8 may suggest resolution limitations.

        \item \textbf{Regions with 1–2 Phases:} Partial volume correction is most effective for such regions.
    \end{itemize}
\end{itemize}

\section{Scalability}
\label{annex:scalability}

When extracting properties from high-resolution samples, memory may present a limiting factor. Commercially available computers often lack sufficient memory to effectively analyze such samples. Furthermore, GPU memory may be lacking or inadequate. To address these issues, we have integrated a Snakemake \cite{molder_sustainable_2021} workflow into our software, which can be executed using our command line interface. Consequently, the software operates in a headless mode and does not require a graphical frontend. Users simply need to provide the path to the grayscale folders, along with their parameters file, to initiate the workflow. As a result, the user receives a project folder that can be transferred to any other device and subsequently loaded using our software for interactive exploration. Prior to this, users can utilize the interactivity of our software on a subset of the data to determine the optimal parameters. For long term reproducibility we provide a container based on docker that has been created using the album docker plugin. We refer to the github page \href{https://gitlab.com/ida-mdc/ari3d.git}{https://gitlab.com/ida-mdc/ari3d.git} for usage instruction.

\section{Example Application}
\label{annex:example}
We exemplified our graphical user interface with a material consists of particles (600-850 um) of a chromite ore. The material has been characterized using MSPaCMAn \cite{gupta_standardized_2024} as composed of only two classes, light material and chromite. Nevertheless, it is known that the light material is composed of quartz as the main phase and alumina silicates with slightly higher attenuation. Here, with the assistance of the GUI those two light classes could be distinguished. It should be noted that finding the appropriate parameters took only a couple of hours and no coding experience as opposed to several days necessary to adjust the script to do the quantification in the original study.  

We further enhanced our software with an artificial dataset for quality control. Within a controlled three-dimensional space measuring 200x200x200 units, we positioned five particles, each with dimensions of 40x40x40 units, composed of one to five artificial material classes. The classes were selected to encompass the full spectrum of the 16-bit gray-values utilized in equivariant distances. Due to the manual creation of these particles, the composition percentages were known in advance. The results of the quantification workflow were tested against this prior knowledge, thus validating our methodology employed in this software. The quality control Jupyter notebook is made available here: \href{https://gitlab.com/ida-mdc/ari3d/-/blob/main/src/ari3d/utils/quality_control.ipynb}{https://gitlab.com/ida-mdc/ari3d/-/blob/main/src/ari3d/utils/quality\_control.ipynb}.

\includegraphics[width=0pt,height=0pt]{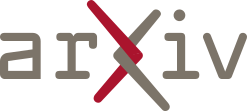}

\end{document}